%% file: neurips_2019.tex
\documentclass{article}

\PassOptionsToPackage{numbers, compress}{natbib}

\usepackage[preprint]{neurips_2019}

\usepackage{float}
\usepackage{graphics,graphicx}
\usepackage{xcolor}
\usepackage{soul}
\usepackage{amsmath,mathtools}
\usepackage{subfigure}

\usepackage[utf8]{inputenc} 
\usepackage[T1]{fontenc}    
\usepackage{hyperref}       
\usepackage{url}            
\usepackage{booktabs}       
\usepackage{amsfonts}       
\usepackage{nicefrac}       
\usepackage{microtype}      
\usepackage[acronym,smallcaps]{glossaries}

\newcommand{\fref}[1]{Fig. \ref{#1}}

\input{default_content/acronyms}

\title{Learning to Play Soccer by Reinforcement and Applying Sim-to-Real to Compete in the Real World}

\author{
   Hansenclever F. Bassani \\
   CIn, Universidade Federal de Pernambuco \\
   Recife, PE, Brazil, 50.740-560 \\
   \texttt{hfb@cin.ufpe.br} \\
   \And
   Renie A. Delgado \\
   CIn, Universidade Federal de Pernambuco \\
   Recife, PE, Brazil, 50.740-560 \\
   \texttt{rad@cin.ufpe.br} \\
   \AND
   José Nilton de O. Lima Junior \\
   CIn, Universidade Federal de Pernambuco \\
   Recife, PE, Brazil, 50.740-560 \\
   \texttt{jnolj@cin.ufpe.br} \\
   \And
   Heitor R. Medeiros \\
   CIn, Universidade Federal de Pernambuco \\
   Recife, PE, Brazil, 50.740-560 \\
   \texttt{hrm@cin.ufpe.br} \\
   \AND
   Pedro H. M. Braga \\
   CIn, Universidade Federal de Pernambuco \\
   Recife, PE, Brazil, 50.740-560 \\
   \texttt{phmb4@cin.ufpe.br} \\
   \And
   Alain Tapp \\
   Mila, Universite de Montréal \\
   Montréal, Québec, Canada, H3C 3J7 \\
   \texttt{alain.tapp@gmail.com} \\
}

\begin{document}

\maketitle


\input{sections/1-introduction.tex}

\input{sections/2-research-problem}
\input{sections/3-motivation}
\input{sections/4-tech-contribution}



\bibliographystyle{abbrvnat}
\bibliography{default_content/bibliography}

\end{document}

%% file: default_content/acronyms.tex
\newacronym{rl}{RL}{\textit{Reinforcement Learning}}
\newacronym{ml}{ML}{\textit{Machine Learning}}
\newacronym{marl}{MARL}{\textit{Multi-agent Reinforcement Learning}} 
\newacronym{il}{IL}{\textit{Independent Learners}} 
\newacronym{jal}{JAL}{\textit{Joint-Action Learners}} 

\newacronym{vsss}{VSSS}{\textit{IEEE Very Small Size Soccer}}
\newacronym{ssl}{SSL}{\textit{Small Size League}}
\newacronym{mdp}{MDP}{\textit{Markov Decision Process}}
\newacronym{larc}{LARC}{\textit{Latin American Robotics Competition}}

\newacronym{dqn}{DQN}{\textit{Deep Q Network}}
\newacronym{ddpg}{DDPG}{\textit{Deep Deterministic Policy Gradient}}

%% file: sections/1-introduction.tex
\section{Introduction}


This work presents an application of \gls{rl} for the complete control of real soccer robots of the \gls{vsss} \cite{vss_rules}, a traditional league in the \gls{larc}. In the \gls{vsss} league, two teams of three small robots play against each other. We propose a simulated environment in which continuous or discrete control policies can be trained, and a Sim-to-Real method to allow using the obtained policies to control a robot in the real world. The results show that the learned policies display a broad repertoire of behaviors which are difficult to specify by hand. This approach, called VSSS-RL, was able to beat the human-designed policy for the striker of the team ranked 3rd place in the 2018 \gls{larc}, in 1-vs-1 matches.

\begin{figure}[ht]
  \centering
  \subfigure[]{\includegraphics[width=0.24\linewidth,scale=1.0]{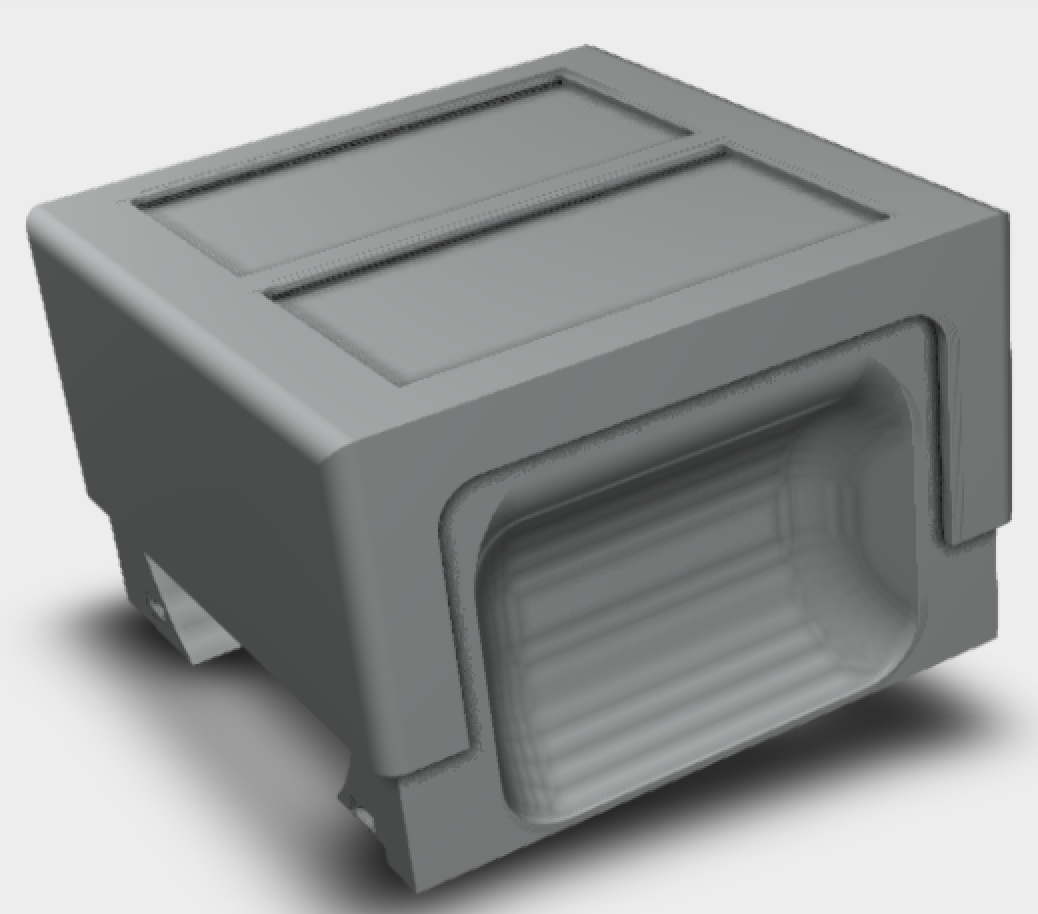}
  \label{fig:robot}}
  \subfigure[]{\includegraphics[width=0.36\linewidth,scale=1]{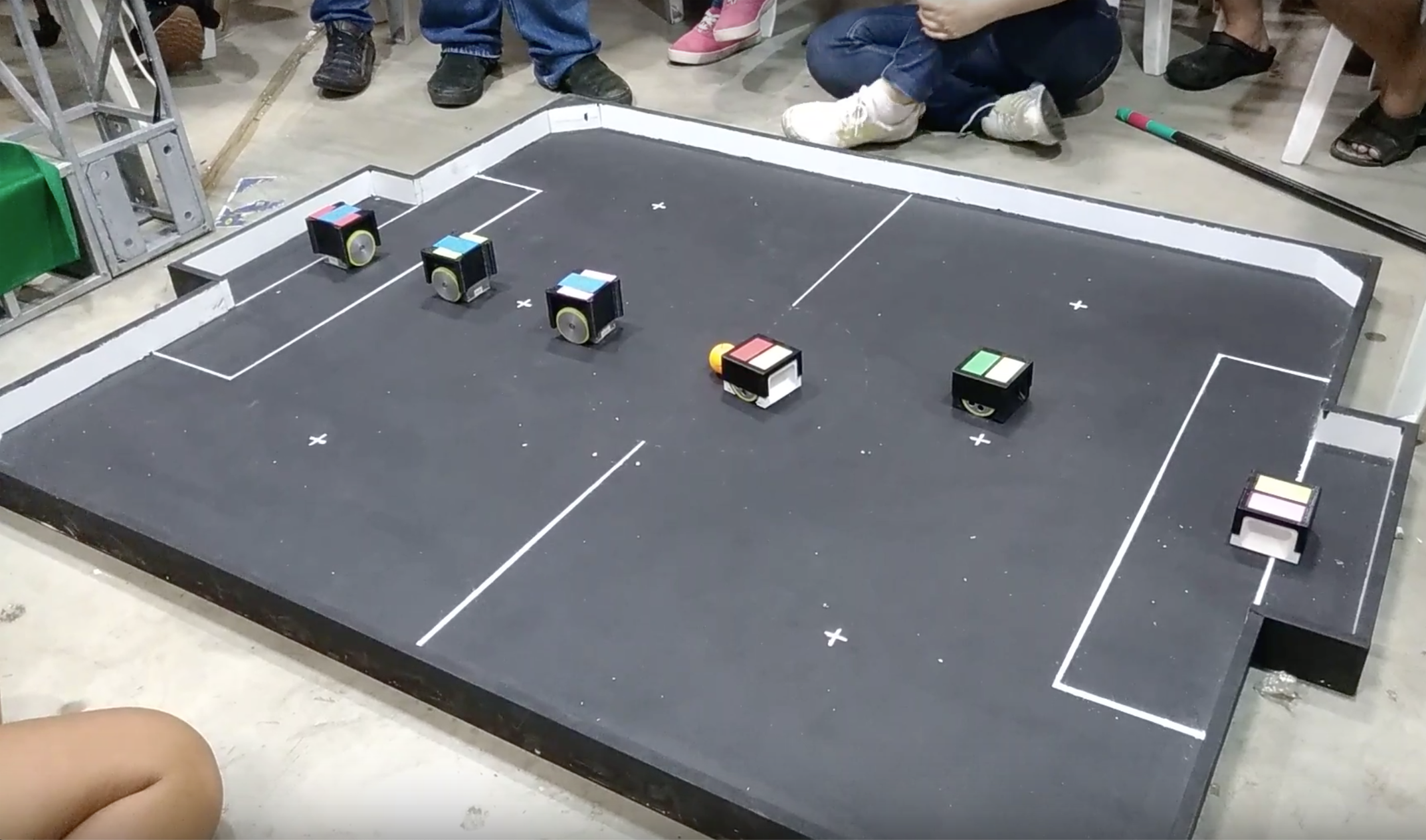}
  \label{fig:vss-real}}
  \subfigure[]{\includegraphics[width=0.36\linewidth,scale=1]{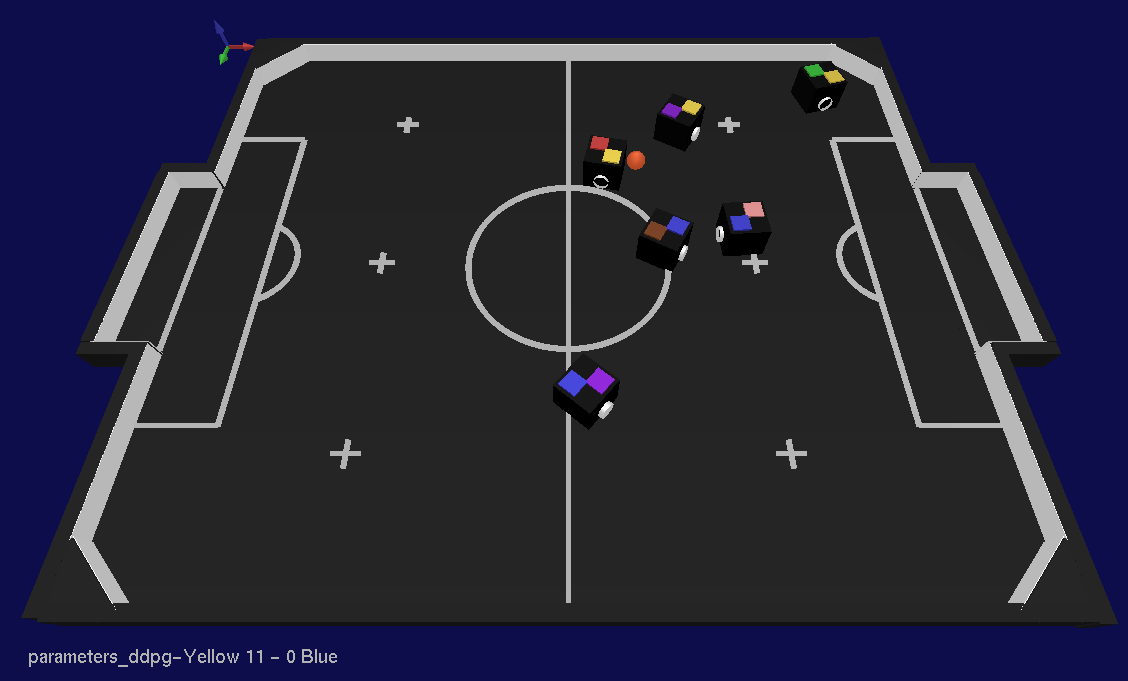}
  \label{fig:vss-sdk}}
  \caption{(a) 3D model of a \gls{vsss} robot; (b) Real-world game setup; and (c) Simulation \cite{vss_sdk}.}
  \label{fig:vss_sim}
\end{figure}

%% file: sections/2-research-problem.tex
\section{Research Problem}

The \gls{vsss} robots are usually programmed to behave adequately in every situation identified by the programmers, employing path planning, collision avoidance, and PID control methods \cite{kim2004soccer}. However, it is extremely hard to foreseen and tackle every possible situation in a dynamic game such as soccer. Therefore, it is clear the need for data-oriented approaches such as \gls{rl}.

However, several barriers exist for applying \gls{rl} successfully in the real world \cite{dulac2019challenges}, as the large amounts of interactions required by the agents to achieve adequate performance are impractical due to degradation of hardware, energy consumption and time required. Thus, the research problem considered in this work is the application of the Sim-to-Real approach, in which the agents are trained in simulation and policies learned are transferred to the real robots.

%% file: sections/3-motivation.tex
\section{Motivation}

Deep \gls{rl} is a suitable approach for learning control and complex behaviors by interacting with the environment since it requires only the specification of a reward function that expresses the desired goals. In the literature of robot soccer, \gls{rl} has been applied for learning specific behaviors, such as kicking \cite{riedmiller2007experiences} and scoring penalty goals \cite{hester2010generalized}. 

Recently, two \gls{rl} soccer simulation environments have been proposed: MuJoCo Soccer \cite{todorov2012mujoco} and Google Research Football \cite{kurach2019google}. However, they are not suitable for the study of Sim-to-Real, because they either do not consider important physical and dynamical aspects or represent a very complex scenario that is not achievable by current robotics technology. Therefore, the need for such an adequate environment, allowing the study of the combination of \gls{rl} with Sim-to-Real in dynamic, multi-agent, competitive, and cooperative situations, is the main motivation behind this work.

%% file: sections/4-tech-contribution.tex
\section{Technical Contribution}

\begin{figure*}[ht!]
  \centering
  \subfigure[]{\includegraphics[width=0.65\textwidth,scale=1]{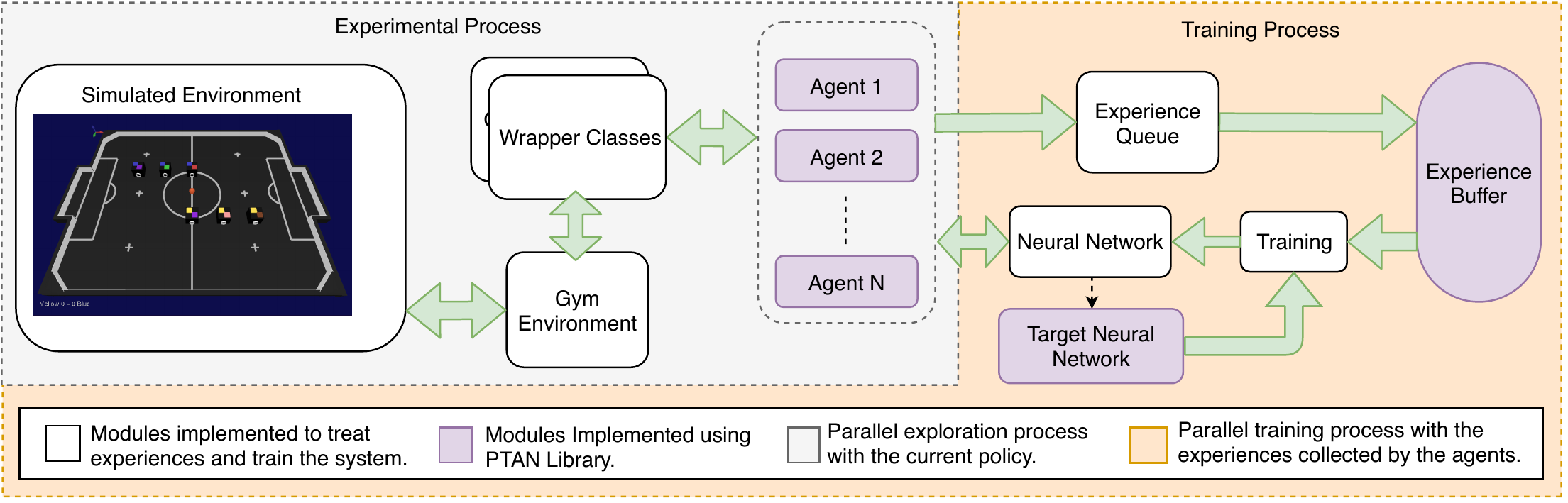}
  \label{fig:architecture}}
  \subfigure[]{\includegraphics[width=0.33\linewidth,scale=1]{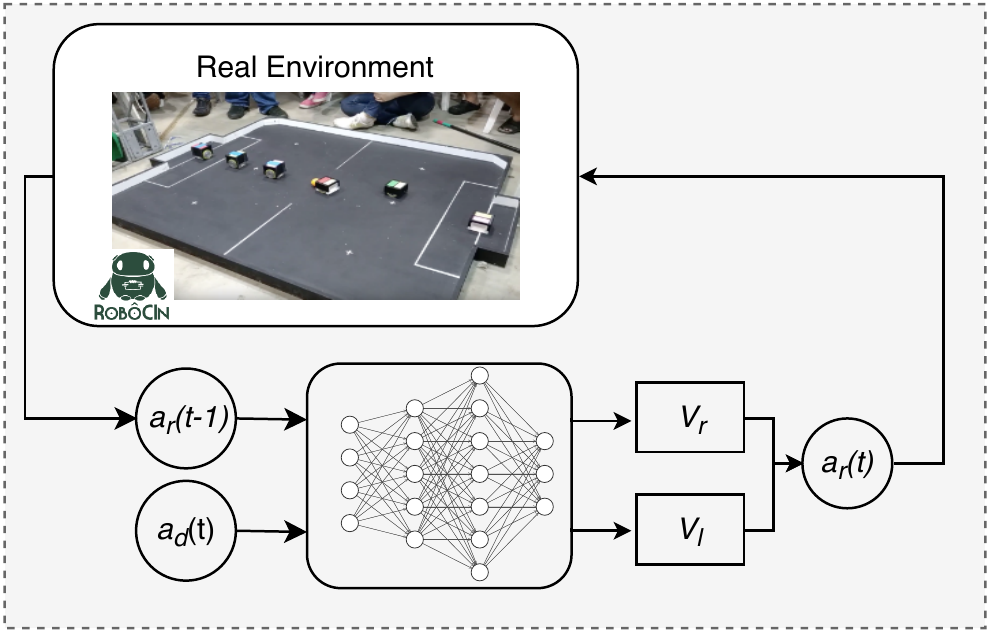}
  \label{fig:sim2realtrain}}

  \caption{VSSS-RL: (a) Environment Architecture for training high-level control policies. (b) Low-level control training processes to enable Sim-to-Real transfer.}
  \label{fig:vssss-env}
\end{figure*}

We propose a simulated environment called \gls{vsss}-RL\footnote{Source code will be available soon at: \url{https://github.com/robocin/vss-environment}}, which supports continuous or discrete control policies. It includes a customized version of the VSS SDK simulator \cite{vss_sdk} and builds a set of wrapper modules to be compatible with the OpenAI Gym standards \cite{gym}. It consists of two main independent processes: the experimental, and the training process. In the first, an OpenAI Gym environment parser was developed, and wrapper classes were implemented to communicate with the agents. In the latter, the collected experiences are stored in an experience buffer that is used to update the policies, as illustrated in \fref{fig:architecture}.

We also proposed a Sim-to-Real method to transfer the obtained policies to a robot in the real world. It is a Domain Adaptation method \cite{andrychowicz2018learning}, consisting of a Feed-Forward Neural Network which learns to map the desired high-level actions $a_{d}(t) = \{v, \omega\}$ (linear and angular speeds) to low-level control commands for the wheel speeds ($V_R$ and $V_L$) (\fref{fig:sim2realtrain}).

\subsection{Experimental Results}

The results, submitted to ICRA2020, show that the two baseline \gls{rl} methods evaluated, \gls{ddpg} \cite{lillicrap2015continuous} and \gls{dqn} \cite{volodymyr2013playing}, were able to learn suitable policies in simulation when applying reward shaping \cite{sutton1998introduction}. The learned polices display rich and complex behaviors\footnote{See the video available at: \url{https://youtu.be/a9dTMtanh-U}} extremely difficult to specify by hand as well as to identify the correct moments when they should be applied. Moreover, the proposed Sim-to-Real method employed allowed us to achieve similar results in the real world in terms of average steps to score a goal ($547.2 \pm 233.6$ in simulation and $456.8 \pm 147.2$ in the real world). 

Finally, the complete approach was evaluated in 1-vs-1 matches against the striker of RoboCIn VSSS team, 3rd place on the LARC 2018. The final scores of the matches were 19 for VSSS-RL and 13 for RoboCIn in the first game, and 22 for VSSS-RL approach and 17 for RoboCIn in the second. These wins highlight the capabilities of the proposed approach.

%% file: neurips_2019.bbl
\begin{thebibliography}{13}
\providecommand{\natexlab}[1]{#1}
\providecommand{\url}[1]{\texttt{#1}}
\expandafter\ifx\csname urlstyle\endcsname\relax
  \providecommand{\doi}[1]{doi: #1}\else
  \providecommand{\doi}{doi: \begingroup \urlstyle{rm}\Url}\fi

\bibitem[vss(2019{\natexlab{a}})]{vss_rules}
{Very Small Size Soccer Rules}.
\newblock
  \url{http://www.cbrobotica.org/wp-content/uploads/2014/03/VerySmall2008\_en.pdf},
  2019{\natexlab{a}}.
\newblock [Online; accessed 26-May-2019].

\bibitem[vss(2019{\natexlab{b}})]{vss_sdk}
{VSS SDK}.
\newblock \url{https://vss-sdk.github.io/book/general.html},
  2019{\natexlab{b}}.
\newblock [Online; accessed 5-June-2019].

\bibitem[Andrychowicz et~al.(2018)Andrychowicz, Baker, Chociej, Jozefowicz,
  McGrew, Pachocki, Petron, Plappert, Powell, Ray,
  et~al.]{andrychowicz2018learning}
M.~Andrychowicz, B.~Baker, M.~Chociej, R.~Jozefowicz, B.~McGrew, J.~Pachocki,
  A.~Petron, M.~Plappert, G.~Powell, A.~Ray, et~al.
\newblock Learning dexterous in-hand manipulation.
\newblock \emph{arXiv preprint arXiv:1808.00177}, 2018.

\bibitem[Brockman et~al.(2016)Brockman, Cheung, Pettersson, Schneider,
  Schulman, Tang, and Zaremba]{gym}
G.~Brockman, V.~Cheung, L.~Pettersson, J.~Schneider, J.~Schulman, J.~Tang, and
  W.~Zaremba.
\newblock Openai gym, 2016.

\bibitem[Dulac-Arnold et~al.(2019)Dulac-Arnold, Mankowitz, and
  Hester]{dulac2019challenges}
G.~Dulac-Arnold, D.~Mankowitz, and T.~Hester.
\newblock Challenges of real-world reinforcement learning.
\newblock \emph{arXiv preprint arXiv:1904.12901}, 2019.

\bibitem[Hester et~al.(2010)Hester, Quinlan, and Stone]{hester2010generalized}
T.~Hester, M.~Quinlan, and P.~Stone.
\newblock Generalized model learning for reinforcement learning on a humanoid
  robot.
\newblock In \emph{2010 IEEE International Conference on Robotics and
  Automation}, pages 2369--2374. IEEE, 2010.

\bibitem[Kim et~al.(2004)Kim, Kim, Kim, and Seow]{kim2004soccer}
J.-H. Kim, D.-H. Kim, Y.-J. Kim, and K.~T. Seow.
\newblock \emph{Soccer robotics}, volume~11.
\newblock Springer Science \& Business Media, 2004.

\bibitem[Kurach et~al.(2019)Kurach, Raichuk, Sta{\'n}czyk, Zajac, Bachem,
  Espeholt, Riquelme, Vincent, Michalski, Bousquet, et~al.]{kurach2019google}
K.~Kurach, A.~Raichuk, P.~Sta{\'n}czyk, M.~Zajac, O.~Bachem, L.~Espeholt,
  C.~Riquelme, D.~Vincent, M.~Michalski, O.~Bousquet, et~al.
\newblock Google research football: A novel reinforcement learning environment.
\newblock \emph{arXiv preprint arXiv:1907.11180}, 2019.

\bibitem[Lillicrap et~al.(2015)Lillicrap, Hunt, Pritzel, Heess, Erez, Tassa,
  Silver, and Wierstra]{lillicrap2015continuous}
T.~P. Lillicrap, J.~J. Hunt, A.~Pritzel, N.~Heess, T.~Erez, Y.~Tassa,
  D.~Silver, and D.~Wierstra.
\newblock Continuous control with deep reinforcement learning.
\newblock \emph{arXiv preprint arXiv:1509.02971}, 2015.

\bibitem[Riedmiller and Gabel(2007)]{riedmiller2007experiences}
M.~Riedmiller and T.~Gabel.
\newblock On experiences in a complex and competitive gaming domain:
  Reinforcement learning meets robocup.
\newblock In \emph{2007 IEEE Symposium on Computational Intelligence and
  Games}, pages 17--23. IEEE, 2007.

\bibitem[Sutton et~al.(1998)Sutton, Barto, et~al.]{sutton1998introduction}
R.~S. Sutton, A.~G. Barto, et~al.
\newblock \emph{Introduction to reinforcement learning}, volume~2.
\newblock MIT press Cambridge, 1998.

\bibitem[Todorov et~al.(2012)Todorov, Erez, and Tassa]{todorov2012mujoco}
E.~Todorov, T.~Erez, and Y.~Tassa.
\newblock Mujoco: A physics engine for model-based control.
\newblock In \emph{2012 IEEE/RSJ International Conference on Intelligent Robots
  and Systems}, pages 5026--5033. IEEE, 2012.

\bibitem[Volodymyr et~al.(2013)Volodymyr, Kavukcuoglu, Silver, Graves, and
  Antonoglou]{volodymyr2013playing}
M.~Volodymyr, K.~Kavukcuoglu, D.~Silver, A.~Graves, and I.~Antonoglou.
\newblock Playing atari with deep reinforcement learning.
\newblock In \emph{NIPS Deep Learning Workshop}, 2013.

\end{thebibliography}
